\documentclass{article} 
\usepackage{iclr2019_conference,times}

\usepackage{amsmath,amsfonts,bm}

\def\eqref#1{equation~\ref{#1}}

\def\1{\bm{1}}

\DeclareMathAlphabet{\mathsfit}{\encodingdefault}{\sfdefault}{m}{sl}
\SetMathAlphabet{\mathsfit}{bold}{\encodingdefault}{\sfdefault}{bx}{n}

\newcommand{\R}{\mathbb{R}}

\usepackage{hyperref}
\usepackage{url}
\usepackage{booktabs}
\usepackage{multirow}
\usepackage{multicol}
\usepackage{appendix}

\usepackage{tikz}
\usetikzlibrary{plotmarks}

\title{Spatio-Temporal Deep Graph Infomax}

\author{Felix L. Opolka\thanks{Equal contribution}, Aaron Solomon\textsuperscript{*}, C\u{a}t\u{a}lina Cangea, Petar Veli\v{c}kovi\'{c}, Pietro Li\`{o} \\
Department of Computer Science and Technology\\
University of Cambridge\\
\texttt{\{flo23,acs207,ccc53,pv273,pl219\}@cam.ac.uk} \\
\And
R Devon Hjelm \\
Microsoft Research \\
Mila -- Qu\'{e}bec Artificial Intelligence Institute \\
\texttt{devon.hjelm@microsoft.com} \\
}

\newcommand{\bd}[1]{\boldsymbol{#1}}

\iclrfinalcopy 
\begin{document}

\maketitle

\begin{abstract}
Spatio-temporal graphs such as traffic networks or gene regulatory systems present challenges for the existing deep learning methods due to the complexity of structural changes over time. To address these issues, we introduce Spatio-Temporal Deep Graph Infomax (STDGI)---a fully unsupervised node representation learning approach based on mutual information maximization that exploits both the temporal and spatial dynamics of the graph. Our model tackles the challenging task of node-level regression by training embeddings to maximize the mutual information between patches of the graph, at any given time step, and between features of the central nodes of patches, in the future. We demonstrate through experiments and qualitative studies that the learned representations can successfully encode relevant information about the input graph and improve the predictive performance of spatio-temporal auto-regressive forecasting models.

\end{abstract}

\section{Introduction and Related Work}

Using deep learning techniques to analyze spatio-temporal data has shown promising results on classification and regression tasks, in contexts such as road and traffic networks~\citep{shi2018}, gene expression data~\citep{Dutil2018}, and Internet networking~\citep{Boutaba2018}. Previous work has leveraged supervised learning techniques, largely by combining auto-regressive models with graph convolutional layers of different types~\citep{yu2017, li2018, zhang2018}. In this work, we address this task in the unsupervised learning setting and propose an approach for learning node representations for a spatio-temporal graph in a fully unsupervised fashion, encoding useful information that increases performance of a downstream forecasting model.

While neural network based learning methods for graph-structured data have received substantial attention, previous work has largely focused on supervised classification tasks. \cite{velickovic2018} have recently proposed Deep Graph Infomax (DGI)---an unsupervised representation learning approach for nodes in non-temporal graphs---and achieved state-of-the-art performance on classification benchmarks. Unlike previous methods~\citep{Mutlu2018}, DGI does not rely on a random walk or adjacency-based methods and instead uses graph convolutions~\citep{kipf2016} to build on the deep mutual information maximization principle described by~\cite{Hjelm2018}. So far, DGI has only been applied to non-temporal graphs in the node classification setting. In this work, we adapt the mutual information maximization principle to spatio-temporal graphs and show that the learned embeddings can encode valuable information for node regression tasks. We compare our model to a baseline auto-regressive model that only exploits temporal information and thus show that we can encode relevant spatial information in a fully unsupervised manner.

\section{Background}

\subsection{Problem Setting}

The node regression task takes the form of a prediction on a graph $G = (V, E, \bd{W})$ with node set $V$, edge set $E$, and weighted adjacency matrix $\bd{W}$. Node features change over time and hence features at time step $t$ are given by $\bd{X}^{(t)}$. Given the features from the most recent $T'$ time steps, the task is to predict the features of the next $T$ time steps using a function $f(\cdot)$, potentially parameterized by a neural network:
    \begin{equation}
        [\bd{X}^{(t-T'+1)}, \cdots, \bd{X}^{(t)}; G] \overset{f(\cdot)}{\xrightarrow{\hspace*{0.5cm}}} [\bd{X}^{(t+1)}, \cdots, \bd{X}^{(t+T)}].
    \end{equation}

\subsection{Learning Representations via Mutual Information Maximization}


Deep InfoMax~\citep[DIM,][]{Hjelm2018} is a recent approach for unsupervised representation learning that derives embeddings by maximizing the mutual information between the output of an encoder and local patches of the input. DIM builds on Mutual Information Neural Information~\citep[MINE,][]{Belghazi2018}, which formulates an estimate $\widehat{I}(X; Y)$ for the mutual information between random variables $X, Y$ using neural networks. These estimates are obtained by training a classifier (a.k.a, the \emph{discriminator} or \emph{statistics network}) to distinguish between samples from the joint distribution and the product of marginals. DIM applies this approach to representation learning by training both the encoder and the discriminator to maximize the mutual information between the random variables corresponding to local input patches and the embeddings. Deep Graph Infomax (DGI) extends this representation learning technique to non-temporal graphs, finding node embeddings that maximize the mutual information between local patches of the graph and summaries of the entire graph. Here, we build on these methods and propose a representation learning technique for spatio-temporal graphs. Furthermore, unlike in previous work, we evaluate our embeddings in the regression rather than classification setting.

\section{Spatio-Temporal Deep Graph Infomax}

We extend the DGI approach by adapting it to spatio-temporal graphs and refer to our method as \textit{spatio-temporal deep graph infomax} (STDGI).
At each time step, representations are trained for each node in the graph in a fully unsupervised fashion. Similarly to DIM, we train the encoder to maximize the mutual information between patches in the graph at a particular time step $t$ and the raw features of the same node at a future time step $t + k$. The goal is to aggregate, for each node, the information from its neighbourhood that is most relevant for predicting its features in the future.



\subsection{Architecture Components}

\begin{figure}
    \centering
    \includegraphics[width=0.8\textwidth,keepaspectratio]{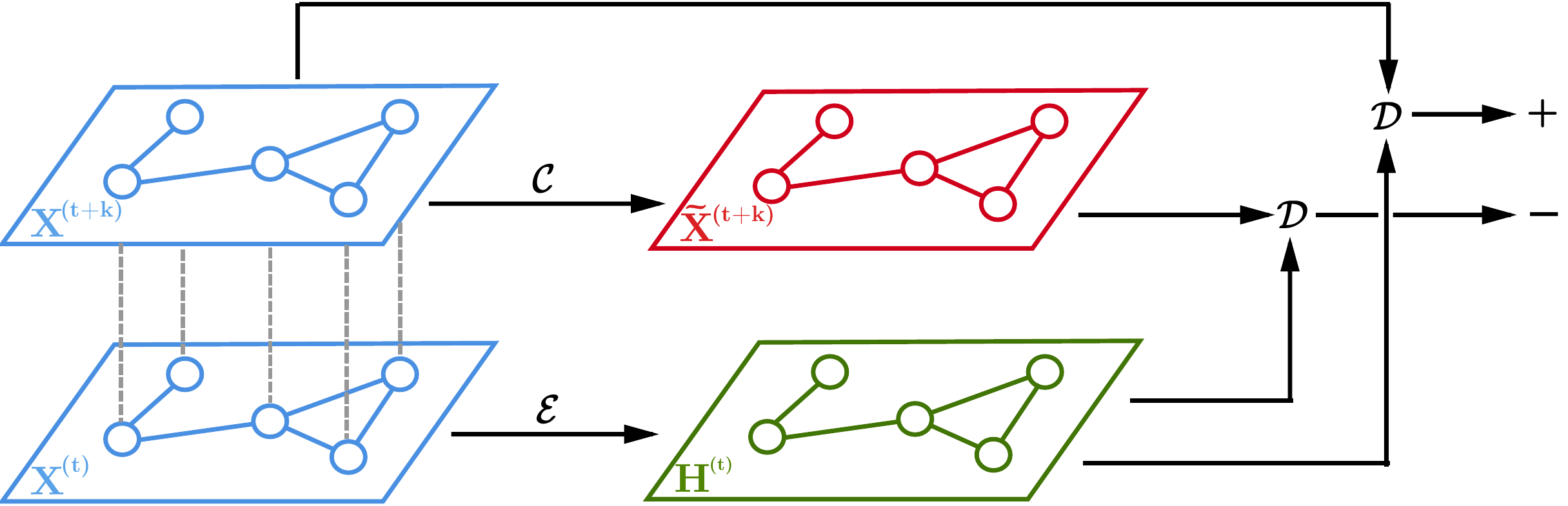}
    \caption{Visualization of the unsupervised training procedure. The embeddings $\bd{H}^{(t)}$ for the nodes at time step $t$ are computed by the encoder $\mathcal{E}$. Given the embeddings and the uncorrupted raw features $\bd{X}^{(t+k)}$, the discriminator $\mathcal{D}$ is trained to output that the sample is positive. When given the embeddings and the corrupted raw features $\bd{\tilde{X}}^{(t+k)}$, the discriminator is trained to output that the sample is negative.}
    \label{fig:architecture}
\end{figure}

The unsupervised training setup is equivalent to the one in DIM or DGI. An encoder $\mathcal{E}$ computes embeddings $\bd{h}_{v}^{(t)} \in \R^{K}$ for each node $v \in V$ at each time step $t$, using node features $\bd{X}$ and the graph structure $\bd{W}$.
The discriminator $\mathcal{D}$ then receives pairs $(\bd{h}_{v}^{(t)}, \bd{x}_{v}^{(t+1)})$ containing the embedding and raw features of the same node at the current and next time step, respectively.
We refer to such a pair of embedding and raw features as a \textit{positive sample}, if both were drawn from the same graph. A \textit{negative sample} will then consist of an embedding and raw features, where the latter are obtained from a corrupted version of the graph, derived by randomly permuting the node features of the graph at each time step. Positive samples can be understood as being drawn from the \emph{joint distribution} of embeddings and raw features, whereas negative samples are drawn from the \emph{marginal distributions}.
The discriminator outputs a score corresponding to whether a given pair represents a positive or negative sample; both the encoder and discriminator are trained jointly to distinguish between positive and negative samples by minimizing the binary cross entropy loss. This maximizes the mutual information between the embeddings and the raw features of the next time step~\citep{poole2018, Belghazi2018, Hjelm2018}.

During supervised training, the embeddings $\bd{h}_{v}^{(t)}$ output by the encoder are concatenated with the raw features $\bd{x}_{v}^{(t)}$. In order to use the learned representations for our considered task, the resulting features $\bd{X}^{*(t)} = \bd{X}^{(t)} \oplus \bd{H}^{(t)}$ serve as input to a \emph{downstream supervised regressor}.

\section{Experiments}

We devise an evaluation setup for the \emph{traffic forecasting task} to determine whether embeddings successfully encode spatial information of the graph that is relevant for making more accurate predictions. From a high-level point of view, the graph structure corresponds to a network of traffic sensors, where nodes are individual traffic sensors. An edge between two nodes is added when the distance between the two corresponding sensors is below a certain threshold. The time series of node features are given by the traffic measurements of each sensor over time.

\subsection{Experimental Setup}

We use the METR-LA dataset~\citep{jagadish2014}, which contains data recorded by 207 traffic sensors. The traffic measurements were aggregated into five minute intervals and consist of traffic speed and the time of day.
The graph is given by a directed, weighted adjacency matrix. The edge weights are the exponentially decaying distances along the roads. For more details on the data set, we refer to~\cite[E.1]{jagadish2014}.
Given the past 12 time steps (corresponding to measurements over 1h), the predictor has to forecast the traffic speeds at the next 12 time steps.


The encoder consists of a linear layer applied to each $\bd{x}_{v}^{(t)}$, followed by two graph convolutional layers applied to each $\bd{X}^{(t)}$. 
The discriminator is a two-layer fully-connected neural network, which concatenates the embedding and raw features of each pair and outputs whether the pair is a positive or negative sample. We chose to train three separate discriminators of the same architecture. Each discriminator compares the embedding to the raw feature of the same node $k$ steps in the future where $k = 1, 3, 6$. 
For the downstream regressor, we employ an LSTM seq2seq model~\citep{sutskever2014sequence}, which operates on the time series of each node in isolation. As a baseline, we compare our regressor to one with an identical configuration that receives as input \emph{only the raw features} (rather than the concatenation of the raw features and embeddings). More details on the experimental setup can be found in Appendix~\ref{sec:hyperparams}.



\subsection{Results}

\begin{table}
    \centering
        \begin{tabular}{ccccc} 
        \toprule
        \textbf{Method} & & \textbf{15 min} & \textbf{30 min} & \textbf{60 min} \\
        \midrule
         & & \multicolumn{3}{c}{\textbf{MAE}} \\
        \midrule
        LSTM Baseline & & $3.67 \pm 0.035$ & $4.88 \pm 0.017$ & $6.53 \pm 0.015$ \\
        STDGI & & $\bd{3.61} \pm 0.0038$ & $\bd{4.80} \pm 0.044$ & $\bd{6.41} \pm 0.016$ \\
        \midrule
         & & \multicolumn{3}{c}{\textbf{RMSE}} \\
        \midrule
        LSTM Baseline & & $8.51 \pm 0.037$ & $10.66 \pm 0.0076$ & $13.04 \pm 0.018$ \\
        STDGI & & $\bd{8.28} \pm 0.0042$ & $\bd{10.42} \pm 0.0055$ & $\bd{12.79} \pm 0.018$ \\
        \midrule
         & & \multicolumn{3}{c}{\textbf{MAPE}} \\
        \midrule
        LSTM Baseline & & $7.8 \pm 0.03\%$ & $9.9 \pm 0.06\%$ & $13.1 \pm 0.09\%$ \\
        STDGI & & $\bd{7.6} \pm 0.01\%$ & $\bd{9.5} \pm 0.01\%$ & $\bd{12.5} \pm 0.04\%$ \\
        \bottomrule
        \end{tabular}
    \vspace{0.1cm}
    \caption{\label{tab:results}Comparison of the LSTM baseline regressor and the STDGI method on METR-LA. We compute the MAE, RMSE, and MAPE of predictions for three different time horizons.}
\end{table}

Results for the LSTM regressor that only uses raw features (baseline) and for the one that uses raw features concatenated with STDGI embeddings are shown in Table~\ref{tab:results}. We find that regressors using STDGI embeddings achieve lower predictions errors for all time horizons considered, and that the improvements \emph{become more pronounced for larger time horizons}. This suggests that STDGI extracts embeddings with useful long-term features that improve upon the future predictive ability of raw data alone.  All performances increases documented here are significant at $\alpha = 0.01$.  

\begin{figure}
    \centering
    \includegraphics[width=0.43\textwidth,keepaspectratio]{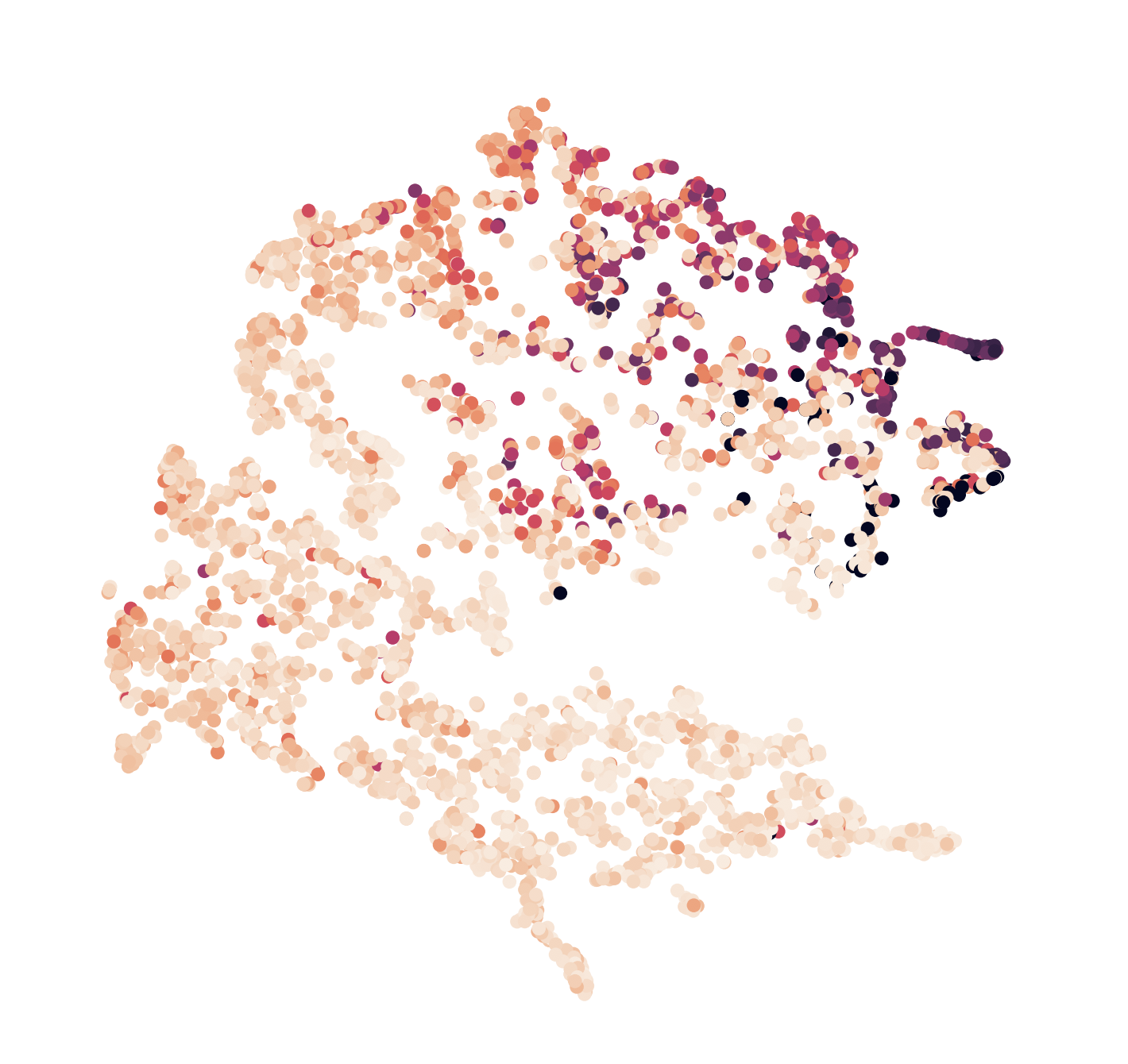}
    \caption{t-SNE visualization of embeddings from randomly selected time points spanning the available time series. Embeddings were visualized in two dimensions via t-SNE and colored by speeds at three timesteps (15 minutes) in the future to demonstrate the relationship between embedding similarity and true speed similarity.}
    \label{fig:tsne}
\end{figure}

These results are further supported by the qualitative study in Figure~\ref{fig:tsne}, which illustrates t-SNE-processed embeddings colored by future time point speeds. As indicated by their color, closely clustered embeddings generally share similar speeds. This suggests that embedding similarity is a proxy for speed similarity, and thus that embeddings are learning useful long-term information.


\section{Conclusion}

We have presented STDGI, an approach for learning embeddings of nodes in a graph that evolves over time, which leverages mutual information maximization and performs node regression in a spatio-temporal prediction context. We demonstrate that an auto-regressive seq2seq model operating on the time axis achieves higher predictive performance when making use of STDGI embeddings, thus confirming that the method successfully encodes valuable information over that provided by the standard baseline, even in the more challenging setting of regression. Moreover, the results show that STDGI is able to capture intrinsic and helpful properties of traffic flow by building increasingly stronger representations of the graph in relation to the baseline, as the prediction time horizon becomes larger. Our model represents both a generalization of DGI to spatio-temporal settings as well as a successful extension of the method to perform regression, in addition to classification. Future work will aim to find embeddings that further increase the accuracy of the downstream regressor and provide high-quality representations for multiple predictive tasks.

\bibliography{iclr2019_conference}
\bibliographystyle{iclr2019_conference}

\begin{appendix}

\section{Details on Experimental Setup}\label{sec:hyperparams}

The METR-LA~\citep{jagadish2014} dataset contains data recorded by 207 traffic sensors throughout Los Angeles County, from March 1st, 2012 to June 20th, 2012.
In our experiments, we use the canonical split of the dataset into training, validation, and test set containing $23974$, $3425$, and $6850$ samples, respectively.

All layers in the encoder contain 64 hidden units and the embeddings size is 128.
The two fully-connected layers of the discriminator contain 6 and 1 hidden units, respectively.
The seq2seq downstream regressor consists of a single LSTM layer~\citep{hochreiter1997} with 64 hidden units.

All models are trained with a batch size of 64 for 120 epochs, using the Adam optimizer~\citep{kingma2014}
The unsupervised training of embeddings is carried out for 100 epochs, with an initial learning rate of $1e^{-3}$ that is reduced by a factor of $\frac{1}{10}$ every 30 epochs after the first 20.
The supervised models use the mean absolute error (MAE) over the entire horizon of 12 steps as the loss function. The learning rate is initially $1e^{-2}$ and decreases by a factor of $\frac{1}{10}$ every 30 epochs after the first 20.

\end{appendix}

\end{document}